\documentclass[10pt,twocolumn,letterpaper]{article}

\usepackage{wacv}
\usepackage{times}
\usepackage{epsfig}
\usepackage{graphicx}
\usepackage{amsmath}
\usepackage{amssymb}
\usepackage{algorithm}
\usepackage{algorithmic}
\usepackage{subfigure}
\usepackage{booktabs}
\usepackage{color}
\usepackage{epstopdf}
\usepackage{xcolor}
\usepackage{tabu}
\usepackage{comment}
\usepackage{multirow}
\usepackage{enumitem}
\usepackage[pagebackref=true,breaklinks=true,letterpaper=true,colorlinks,bookmarks=false]{hyperref}
\usepackage[toc,page]{appendix}

\makeatletter
\DeclareRobustCommand\onedot{\futurelet\@let@token\bmv@onedotaux}
\def\bmv@onedotaux{\ifx\@let@token.\else.\null\fi\xspace}
\def\eg{\emph{e.g}\onedot} 
\def\ie{\emph{i.e}\onedot}

\def\etal{\emph{et al}\onedot}

\makeatother

\newcommand{\mI}{{\boldsymbol{I}}}
\newcommand{\mIh}{{\widehat{\boldsymbol{I}}}}
\newcommand{\mTheta}{{\boldsymbol{\Theta}}}
\newcommand{\mPhi}{{\boldsymbol{\Phi}}}
\newcommand{\vpsi}{{\boldsymbol{\psi}}}
\newcommand{\mG}{{\boldsymbol{G}}}
\newcommand{\mD}{{\boldsymbol{D}}}
\newcommand{\va}{{\boldsymbol{a}}}

\wacvfinalcopy 

\def\wacvPaperID{628} 

\begin{document}

\title{Recovering Faces from Portraits with Auxiliary Facial Attributes}

\author{Fatemeh~Shiri \\
Institution1\\
{\tt\small firstauthor@i1.org}
\and
Second Author \\
Institution2\\
{\tt\small secondauthor@i2.org}
}

\author{Fatemeh~Shiri\textsuperscript{1}, Xin~Yu\textsuperscript{1}, Fatih~Porikli\textsuperscript{1}, Richard~Hartley\textsuperscript{1,2}, Piotr~Koniusz\textsuperscript{2,1,}\thanks{This work is accepted by WACV'19. Our code and model will be released on {\fontsize{9}{9}\selectfont \color{red}{\url{http://claret.wikidot.com}}}.}\\
$^1\!$Australian National University, $^2$Data61/CSIRO\\
{\tt\small firstname.lastname@\{anu.edu.au\textsuperscript{1}, data61.csiro.au\textsuperscript{2}\}}
}

\maketitle

\vspace{-1.4em}
\begin{abstract}
\vspace{-1mm}
Recovering a photorealistic face from an artistic portrait is a challenging task since crucial facial details are often distorted or completely lost in artistic compositions. To handle this loss, we propose an Attribute-guided Face Recovery from Portraits (AFRP) that utilizes a Face Recovery Network (FRN) and a Discriminative Network (DN). FRN consists of an autoencoder with residual block-embedded skip-connections and incorporates facial attribute vectors into the feature maps of input portraits at the bottleneck of the autoencoder.  DN has multiple convolutional and fully-connected layers, and its role is to enforce FRN to generate authentic face images with corresponding facial attributes dictated by the input attribute vectors. 
For the preservation of identities, we impose the recovered and ground-truth faces to share similar visual features. Specifically, DN determines whether the recovered image looks like a real face and checks if the facial attributes extracted from the recovered image are consistent with given attributes. 
Our method can recover photorealistic identity-preserving faces with desired attributes from unseen stylized portraits, artistic paintings, and hand-drawn sketches. On large-scale synthesized and sketch datasets, we demonstrate that our face recovery method achieves state-of-the-art results.
\end{abstract}
\vspace{-1.6em}
\section{Introduction}
\label{sec:introduction}
\begin{figure*}[t]
\begin{center}
\subfigure[Original]{\label{fig:opena}\scalebox{1}[1]{\includegraphics[width=0.13\linewidth]{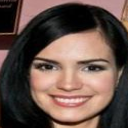}}}
\centering
\subfigure[Portrait]{\label{fig:openb}\scalebox{1}[1]{\includegraphics[width=0.13\linewidth]{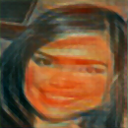}}}
\subfigure[\scriptsize{Landmarks}]{\label{fig:openc}\scalebox{1}[1]{\includegraphics[width=0.13\linewidth]{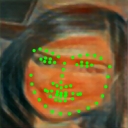}}}
\subfigure[Shiri~\cite{shiri2017face}]{\label{fig:opend}\scalebox{1}[1]{\includegraphics[width=0.13\linewidth]{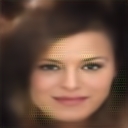}}}
\subfigure[Shiri\cite{shiri2018identity}]{\label{fig:opene}\scalebox{1}[1]{\includegraphics[width=0.13\linewidth]{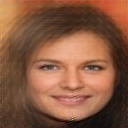}}}
\subfigure[\scriptsize{Pix2Pix~\cite{isola2016image}}]{\label{fig:openf}\scalebox{1}[1]{\includegraphics[width=0.13\linewidth]{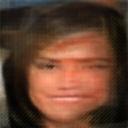}}}
\subfigure[Ours]{\label{fig:openg}\scalebox{1}[1]{\includegraphics[width=0.13\linewidth]{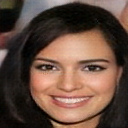}}}
\vspace{-0.2em}
\caption{Comparisons to the state-of-the-art methods. (a) Ground-truth face image (from test dataset; not used in the training). (b) Unaligned stylized portraits of (a) from \emph{Scream} style (unseen style in training), respectively. (c) Detected landmarks by approach~\cite{zhang2014facial}. (d) Results obtained by~\cite{shiri2017face}. (e) Results obtained by \cite{shiri2018identity}. (f) Results obtained by~\cite{isola2016image} (pix2pix). (g) Our results.} 
\label{fig:open}
\vspace{-2.2em}
\end{center}
\end{figure*}

\vspace{-1mm}
Numerous style transfer methods have been proposed to transfer arbitrary artwork styles into content images. In contrast to image stylization, we address a challenging inverse problem called photorealistic face recovery from stylized portraits which aims at recovering a photorealistic face image from a given stylized portrait.
The recovery of the latent photorealistic face from its artistic portrait can help facial analysis and the digital entertainment. 
Facial details in stylized portraits contain artistic effects and distortions such as profile edges and texture changes as shown in Fig.~\ref{fig:openb}. These artistic effects result in a partial loss of facial details and identity-related information.
Moreover, stylized face images may contain various facial expressions, facial distortions and misalignments. Off-the-shelf facial landmark detectors often fail to localize facial landmarks correctly as shown in Fig.~\ref{fig:openc}. Thus, restoring high-quality photorealistic faces from 
artistic portraits is challenging. 

Motivated by such challenges, the recovery of photorealistic images from portraits has recently received some attention  \cite{shiri2017face,shiri2018identity,isola2016image,CycleGAN2017}. 
The existing methods~\cite{shiri2017face,shiri2018identity,isola2016image,CycleGAN2017} take a portrait image 
and then use an autoencoder to generate a photorealistic face image. These methods do not utilize the valuable semantic information in the process of face recovery. Despite training on large-scale datasets, they fail to provide consistent mappings between Stylized Portraits (SP) and ground-truth Real Faces (RF). Thus, they cannot preserve or enforce desired facial attributes in the recovered images.
As shown in Fig. \ref{fig:opend}, \ref{fig:opene} and \ref{fig:openf}, the facial details recovered by the state-of-the-art methods~\cite{shiri2017face,shiri2018identity,isola2016image} are semantically and perceptually inconsistent with the ground-truth images. Inaccuracies range from an unnatural blur to attribute mismatches which include (but are not limited to) \emph{Black Hair} and \emph{Open Mouth}.

Unlike previous works, we propose to use facial attributes as high-level semantic information to boost the quality of recovered face images. Simply embedding the binary facial attribute vector as an additional input channel to the network results in visible distortions (see Fig.~\ref{fig:ablatione}).
We observe that only low-frequency facial components are visible in the stylized input faces as a residual image (the difference between the RF image and the recovered face image) contains the missing high-frequency details. Thus, to recover the high-frequency facial details, we incorporate auxiliary facial attributes into the residual features. 

Based on our observations above, we present a novel Face Recovery Network (FRN) that can use facial attributes during the recovery step. 
Our FRN uses an autoencoder with residual block-embedded skip connections to incorporate visual features obtained from portraits as well as semantic cues provided by facial attributes. 
FRN progressively upsamples the concatenated feature maps through its deconvolutional layers. Moreover, we employ a discriminative network that examines whether a recovered face image resembles an authentic face image and whether the attributes extracted from the recovered face are consistent with the input attributes. 
As a result, our discriminative network can guide the generative network to incorporate semantic information into the recovery process. 
As shown in Fig.~\ref{fig:openg}, our network learns consistent mappings between SP and RF facial patterns and preserves low-frequency details. 
Thus, we can generate realistic face images with details of the ground truth faces (e.g. \emph{Black Hair, Smiling, Straight Hair, Wearing lip stick, pink cheeks}), as in Fig.~\ref{fig:openg}. 

We require a large number of pairs of Stylized Portraits (SP) and Real Face (RF) for training. Thus, we synthesize a large-scale training dataset. However, the choices of styles are numerous--we cannot generate all possible stylized faces for training. To select distinctive styles for training, we use a style-distance metric to measure the style distinctiveness.
For this purpose, we use the Gram matrices 
~\cite{gatys2016image} 
and the Log-Euclidean distance 
~\cite{jayasumana2013kernel} although other non-Euclidean distances can also be  explored \cite{koniusz2018deeper,me_museum}. 
Specifically, we first measure the distance between Gram matrices of stylized images and the average Gram matrix of real faces, and then select the most distinctive styles, \ie~largest distance, for training. 
Furthermore, we note that our CNN filters learned from the data of seen styles (used for the training phase) can also extract informative features from images belonging to unseen styles. Thus, the facial information of unseen stylized portraits can be extracted and used to generate realistic faces, as later demonstrated in our experiments. The main contributions of our work can be summarized as follows:
\renewcommand{\labelenumi}{\Roman{enumi}.}
\vspace{-2mm}
\hspace{-1cm}
\begin{enumerate}[leftmargin=0.7cm]
\item We design a novel framework which removes styles from unaligned stylized portraits. Our framework encodes stylized images with facial attributes and then recovers realistic faces from  encoded feature maps.
\item We propose an autoencoder with residual block-embedded skip-connections to extract residual feature maps from SP inputs and combine the extracted feature maps with facial attributes. In this fashion, we fuse visual and semantic information for best visual results.
\item By manipulating input attribute vectors, our network can generate the realistic faces with desired attributes.
%
%
\end{enumerate}

\vspace*{-0.05in}
To the best of our knowledge, we are the first to use facial attributes for the face recovery from stylized portraits. 

\begin{figure*}[!t]
\centering
\scalebox{1}[0.95]{\includegraphics[width=0.8\linewidth]{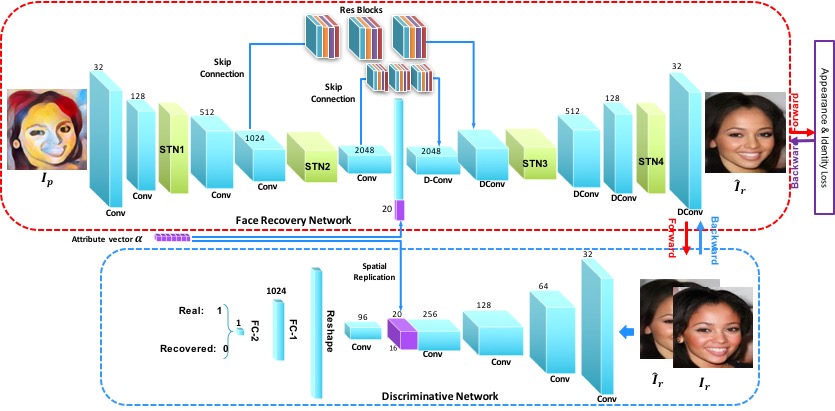}}
\vspace{-0.2em}
\caption{Our attribute-embedded face recovery framework has a generative network (red frame) and a discriminative network (blue frame).}
\label{fig:pipeline}
\vspace{-1.0em}
\end{figure*}
\vspace{-1mm}
\section{Related Work}
\vspace{-2mm}
Below we review papers related to neural style transfer  and deep generative models for image generation.
\vspace{-1mm}
\subsection{Deep Generative Models}
\vspace{-2mm}
Recently, Generative Adversarial Networks (GANs) \cite{Goodfellow2014} have led to large improvements in image generation tasks. 
%
GANs learn the distribution of the training data in a non-conditional setting. Although these methods produce impressive photorealistic images, they cannot distinguish the identities of subjects. Recently, conditional GANs \cite{isola2016image} were used to generate images conditioned on certain input variables. Conditional GANs benefit many applications such as super-resolution \cite{yu2017face,yu2017hallucinating,ledig2017photo,yu2018face}, image generation \cite{oord2016pixel,kingma2014auto,denton2015deep,zhang2017image,shiri2018identity}, image inpainting \cite{yeh2016semantic,pathak2016context}, general purpose image-to-image translation \cite{isola2016image}, image manipulation \cite{zhu2016generative}, synthesize faces
from the landmarks \cite{di2017gp}, and style transfer \cite{ulyanov2016texture}. 
In particular, Li and Wand \cite{li2016precomputed} train a Markovian GAN for the style transfer 
via Markovian neural patches which capture local style statistics. 
Isola \emph{et al.} \cite{isola2016image} develop ``pix2pix'' framework which uses the patch-GAN to transfer low-level features from the input to the output domain. When these patch-based approaches are used to destylize portraits, they produce visual artefacts and fail to capture the global structure of the faces. 

 Sketch from/to photograph synthesis is explored in \cite{nejati2011study,yuen2007human,tang2003face,sharma2011bypassing,sangkloy2017scribbler,wang2018high}. When compared to sketch-to-face synthesis, viewed as a specific case of face recovery, our unified framework is able to process much more complex styles. 

Recently, Yan \etal~\cite{yan2016attribute2image} used a conditional CNN to generate faces based on attributes. Perarnau \etal~\cite{perarnau2016invertible} developped an invertible conditional GAN to generate new faces by editing facial attributes of input images, while Shen and Liu \cite{shen2016learning} manipulated attributes of an input image via its residual image. As their methods are dedicated to generating new face images rather than the face recovery, they cannot preserve identity. 
In contrast, our method uses attributes to reduce the uncertainty of the face recovery and recover faithful realistic faces from artistic portraits. 
\vspace{-1mm}
\subsection{Neural Style Transfer}
\vspace{-2mm}
Style transfer aims to synthesize an image with the visual content of the input image and  a chosen style. 
The seminal work by \cite{gatys2015texture} shows that correlation between feature maps (\ie, Gram matrix 
) captures visual styles. Since then, many follow-up works used 
Gram-based objectives, such as iterative optimization methods \cite{gatys2016image,gatys2017controlling,li2016combining,wilmot2017stable} and feed-forward networks \cite{ulyanov2016texture,johnson2016perceptual,li2016precomputed} which are computationally costly due to the optimization step  at the testing stage. In contrast, feed-forward methods learn a transformation to perform stylization in a feed-forward manner. 

Johnson \emph{et al.} \cite{johnson2016perceptual} train a generative network for a fast style transfer via perceptual loss functions. Their generator network follows \cite{radford2015unsupervised} and uses residual blocks. Texture Network \cite{ulyanov2016texture} uses a multi-resolution generator network. 
Ulyanov \emph{et al.} \cite{ulyanov2017improved} replace the spatial batch norm. with the instance normalization for faster convergence. Wang \emph{et al.}~\cite{wang2017multimodal} enhance the granularity of the feed-forward style transfer via a multimodal CNN and  stylisation losses.

These feed-forward methods are limited by the need to train one network per style due to the lack of generalization in network design. To deal with this restriction, recent approaches encode multiple styles within a single feed-forward network \cite{dumoulin2016,chen2017stylebank,li2017diversified,li2017universal}.
Dumoulin \emph{et al.} \cite{dumoulin2016} use conditional instance normalization to learn necessary normalization parameters for each style. Given feature activations of the content and style images, \cite{chen2016fast} replaces content features with the style features patch-by-patch. 
To achieve an arbitrary style transfer, Chen \emph{et al.}~\cite{chen2017stylebank} swap content features with  style features locally. 
Li~\emph{et al.}~\cite{li2017diversified} adapt a single feed-forward network via a texture controller module. 

As pointed by~\cite{shiri2018identity}, direct use of neural style transfer for face recovery is suboptimal. Even though recent works~\cite{shiri2017face,shiri2018identity} are designed to destylize portrait images, they distort facial details and fail to recover facial traits (\eg, hair color, lipstick, open/closed lips) to match the ground-truth. 
Since  facial traits, such as hair color, are difficult to  infer, the pixel-wise $\ell_2$ norm and perceptual losses cannot yield correct facial attributes. Thus, state-of-the-art face destylization methods produce ambiguous results. 

\vspace{-1mm}
\section{Proposed Method}
\vspace{-1mm}
Below we present an attribute-guided framework for face recovery that takes SP images and facial attribute vectors as inputs, and it outputs photorealistic images of faces. 
\vspace{-2mm}

\begin{figure}[t]
\centering
\includegraphics[width=0.13\linewidth]{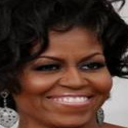}
\includegraphics[width=0.13\linewidth]{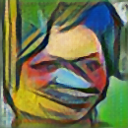}
\includegraphics[width=0.13\linewidth]{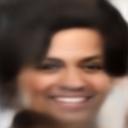}
\includegraphics[width=0.13\linewidth]{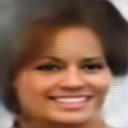}
\includegraphics[width=0.13\linewidth]{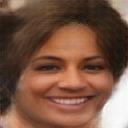}
\includegraphics[width=0.13\linewidth]{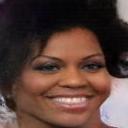}\\
\vspace{-1.2mm}
\subfigure[]{\label{fig:Desa}\scalebox{1}[1]{\includegraphics[width=0.13\linewidth]{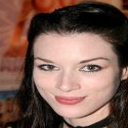}}}
\subfigure[]{\label{fig:Desb}\scalebox{1}[1]{\includegraphics[width=0.13\linewidth]{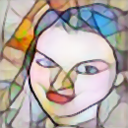}}}
\subfigure[]{\label{fig:Desc}\scalebox{1}[1]{\includegraphics[width=0.13\linewidth]{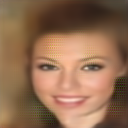}}}
\subfigure[]{\label{fig:Desd}\scalebox{1}[1]{\includegraphics[width=0.13\linewidth]{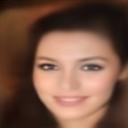}}}
\subfigure[]{\label{fig:Dese}\scalebox{1}[1]{\includegraphics[width=0.13\linewidth]{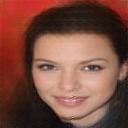}}}
\subfigure[]{\label{fig:Desf}\scalebox{1}[1]{\includegraphics[width=0.13\linewidth]{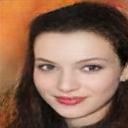}}}
\vspace{-0.2em}
\caption{Contribution of each loss of AFRP network. (a) Ground-truth face images. (b) Input unaligned portraits from unseen styles. (c) Recovered faces without DN or identity-preserving loss. (d) Recovered faces with the $\ell_2$ loss and discriminative loss. (e) Recovered faces with the $\ell_2$, discriminative  and identity-preserving losses. (f) Our final results by embedding facial attributes.}
\label{fig:Effect}
\vspace{-1.5em}
\end{figure}

\subsection{Network Architecture}
\vspace{-2mm}
Our network consists of two parts: a Face Recover Network (FRN) and a Discriminative Network (DN). FRN is composed of an autoencoder with skip connections with residual blocks. 
FRN extracts residual feature maps from input portraits and concatenates the corresponding 20-dimensional attribute vector with the extracted residual feature vector at the bottleneck of the autoencoder and then upsamples it. In this manner, we fuse visual and semantic information to attain high-quality visual performance.
The role of DN is to guide the input attributes 
and the recovered face images to be similar to their real counterparts. The attribute vector is replicated and then concatenated with the extracted feature maps of the convolutional layer of DN.
The entire architecture of our network is shown in Fig.~\ref{fig:pipeline}.

\begin{figure*}[t]
\vspace{-0.4cm}
\begin{center}
\subfigure[]{\label{fig:ablationa}\scalebox{1}[1]{\includegraphics[width=0.11\linewidth]{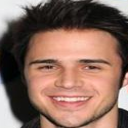}}}
\subfigure[]{\label{fig:ablationb}\scalebox{1}[1]{\includegraphics[width=0.11\linewidth]{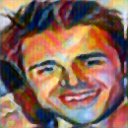}}}
\subfigure[]{\label{fig:ablationc}\scalebox{1}[1]{\includegraphics[width=0.11\linewidth]{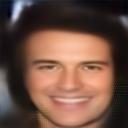}}}
\subfigure[]{\label{fig:ablationd}\scalebox{1}[1]{\includegraphics[width=0.11\linewidth]{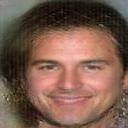}}}\subfigure[]{\label{fig:ablatione}\scalebox{1}[1]{\includegraphics[width=0.11\linewidth]{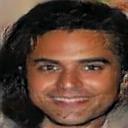}}}
\subfigure[]{\label{fig:ablationf}\scalebox{1}[1]{\includegraphics[width=0.11\linewidth]{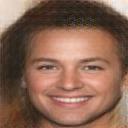}}}
\subfigure[]{\label{fig:ablationg}\scalebox{1}[1]{\includegraphics[width=0.11\linewidth]{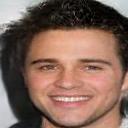}}}
\vspace{-0.2em}
\caption{Ablation study. (a) RF ground-truth image. (b) Unaligned input portrait. (c) Result without  skip connections/residual blocks in the autoencoder. (d) Result without residual blocks in the autoencoder. (e) Result when the attribute vector is
concatenated with the SF input directly. (f) Result without attribute embedding. A standard discriminative network similar to the decoder in~\cite{shiri2018identity}. (g) Our final result.}
\label{fig:ablation}
\end{center}
\vspace{-2em}
\end{figure*}

\noindent{\bf{FRN:}} This module employs a deep fully convolutional autoencoder for face recovery from portraits (see the red frame of Fig.~\ref{fig:pipeline}). The convolutional layers of the encoder capture feature maps of input portraits while deconvolutional layers of the decoder upsample these feature maps to recover facial details.
Previous works~\cite{shiri2017face,shiri2018identity,isola2016image,CycleGAN2017} 
do not use valuable semantic information during face recovery. 
In contrast, our FRN incorporates low-level visual and high-level semantic information (\ie facial attributes) for face recovery to reduce the ambiguity of mappings between SP and RF images. 
Specifically, at the bottleneck of the autoencoder, the attribute vector is concatenated with the residual feature vector (see the purple blocks in Fig.~\ref{fig:pipeline}). A naive embedding of a semantic vector into SP may cause artefacts, \eg Fig.~\ref{fig:ablatione} (encoding input portraits with attributes instead of residual feature maps) 
shows the identity is confused.

We also link symmetrically top convolutional and deconvolutional layers via skip-layer connections~\cite{long2015fully} as they pass higher-resolution visual details of portraits from convolutional to deconvolutional layers for better restoration quality. 
Each skip-connection comprises three residual blocks 
which help our network remove the styles of input portraits while increasing accuracy as in  Fig.~\ref{fig:ablationg}. Without skip-connections, we obtain blurred faces as in Fig.~\ref{fig:ablationc}.

Note that input portraits are misaligned (in-plane rotations, translations). 
Similar to \cite{shiri2018identity}, we use multiple Spatial Transformer Networks (STNs)~\cite{jaderberg2015spatial} in FRN (see the green blocks in Fig.~\ref{fig:pipeline}) 
which compensate for misalignments of input portraits. Thus, our method does not require the use of facial landmarks or 3D face models. 

To measure the appearance similarity between the recovered faces and their RF ground-truth counterparts, we use a pixel-wise $\ell_2$ loss and an identity-preserving loss~\cite{shiri2018identity}. 
The pixel-wise $\ell_2$ loss enforces intensity-based similarity between images of recovered faces and their ground-truth images. The autoencoder supervised by the $\ell_2$ loss tends to output over-smoothed results as shown in Fig.~\ref{fig:Desc}.
For the identity-preserving loss, we use FaceNet~\cite{schroff2015facenet} to extract features from images (see Sec.~\ref{sec:training} for more details), and then we compare the Euclidean distance between features of two images. In this way, we encourage feature similarity between the recovered faces and their ground-truth counterparts. 
Without the identity-preserving loss, the network produces random artefacts that resemble facial details, such as wrinkles, as shown in Fig.~\ref{fig:Desd}.

\noindent{\bf{DN:}} In order to force the FRN to encode facial attribute information, we employ a conditional discriminative network. In particular, the discriminative network distinguishes whether the attributes of face images recovered by FRN match the desired attributes. DN also helps the recovered images to be similar to RF images.
Since our FRN network may learn to ignore attribute vectors, \eg, the weights corresponding to the semantic information are all zero, we design a discriminator network that incorporates semantic attribute information into the generative process.
As shown in the first row of Fig.~\ref{fig:ablationf}, the recovered hair color in the image is brown even if the ground truth hair color is black. This implies that attribute cues are not exploited by the network. Thus, we design a discriminative network which uses attribute embedding in the learning process. 

As shown in the blue frame of Fig.~\ref{fig:pipeline}, DN consists of convolutional and fully connected layers. The real and recovered faces are fed into the network. The attribute information is fed into the middle layer of the network as a conditional information. 
As CNN filters in the first layers extract low-level features and filters in the higher layers extract semantically meaningful image patterns~\cite{zeiler2014visualizing}, in our experiment concatenating features maps with the attribute vectors in the fourth convolutional layer in DN yields better empirical results. 
When there is a mismatch between the extracted features and the input attributes, the discriminative network will pass the errors to the FRN network during backpropagation. 
As shown in Fig.~\ref{fig:ablationg}, our final result matches the ground-truth facial expression, age and gender. 

\vspace{-1mm}
\subsection{Training Procedure}
\vspace{-1mm}
\label{sec:training}
To train AFRP end-to-end, we construct SP, RF and attribute vector triplets $(\mI_p,\mI_r,\va)$
as our training dataset, where $\mI_r$ is the aligned real face image, and $\mI_p$ is the corresponding synthesized unaligned portrait image. 
For each RF, we synthesize different unaligned SP images from various artistic styles to obtain SP/RF training pairs. As detailed in Section~\ref{Sec:Data}, we only use stylized portraits from three distinct styles for training. We use SP image $\mI_p$ and its ground-truth attribute label vector $\va$ as inputs and the corresponding RF ground-truth image $\mI_r$ as a target during training.

We train our FRN network using a pixel-wise $\ell_2$ loss, a feature-wise loss and an adversarial loss to force the generated face $\mI_r$ to resemble its corresponding ground-truth. In addition, we employ a binary cross-entropy loss to update our discriminative network. Since the STN layers are interwoven with the layers of our autoencoder, we optimize the parameters of the autoencoder and the STN layers simultaneously.
Below we explain each loss individually.

\noindent{\bf{Pixel-wise Intensity Similarity Loss: }}
We train a feed-forward network to produce an aligned photorealistic face from a given unaligned portrait. To this end, we feed our FRN with $\mI_p$ images and their corresponding attributes $\va$ as inputs and then force the recovered face $\mIh_r$ to be similar in intensity to its ground-truth counterpart $\mI_r$. Hence, we minimize the objective function $\mathcal{L}_{\small{pix}}$:
\vspace{-2mm}
{
\begin{align}
\label{eqn:pix}
&\fontsize{9.9}{9}\selectfont\!\!\!\!\!\!\!\!\!\!\mathcal{L}_{\small{pix}}(\mTheta)\!=\!\mathbb{E}\|\mIh_r\!-\mI_r\|_F^2\!=\nonumber\\
&\mathbb{E}_{(\mI_p,\mI_r,\va)\sim p(\mI_p,\mI_r,\va)} \|\mG_{\mTheta}(\mI_p,\va) - \mI_r\|_F^2,
\vspace{-1mm}
\end{align}
}
where $\mG_\mTheta(\mI_p,\va)$ and $\mTheta$ represent the output and parameters of our FRN, respectively. We denote $p(\mI_p,\mI_r,\va)$ as the joint distribution of the SP and RF images and the corresponding attributes in the training dataset.

\noindent{\bf{Identity-preserving Loss: }}
To obtain good identity-preserving results, we extract feature maps from the ReLU activations of the FaceNet and compute the Euclidean distance between  features of the recovered face $\mIh_r= \mG_{\mTheta}(\mI_p,\va)$ and ground-truth face $\mI_r$.
As the FaceNet network is pre-trained on a large image dataset, it captures visually meaningful facial features and helps to preserve the identity information. 
Our identity-preserving loss $\mathcal{L}_{id}$ is
\vspace{-1mm}
{
\begin{align}
\label{equ:Id}
&\fontsize{9}{9}\selectfont\!\!\!\!\!\!\!\!\!\!\mathcal{L}_{id}(\mTheta)\!=\!\mathbb{E}\|\vpsi(\mIh_r)\!-\!\vpsi(\mI_r)\|_F^2\!=\nonumber\\
&\fontsize{9}{9}\selectfont\mathbb{E}_{(\mI_p,\mI_r,\va)\sim p(\mI_s,\mI_r,\va)}\|\vpsi(\mG_{\mTheta}(\mI_p,\va))-\vpsi(\mI_r)\|^2_{F} ,
\vspace{-3mm}
\end{align}
}%
where $\vpsi(\cdot)$ denotes the feature maps extracted from the layer ReLU3-2 of the FaceNet.

\noindent{\bf{Discriminative Loss: }}
The discriminative network should note if recovered faces contain the desired attributes and  distinguish recovered faces from real ones. Moreover, FRN should make the discriminative network $\mD_\mPhi$ fail to distinguish recovered faces from real ones and the attributes of generated faces should match the input attributes.
Hence, in order to train the discriminative network, we take real FR face images $\mI_r$ and their corresponding ground-truth attributes $\va$ as positive sample pairs $(\mI_r, \va)$. 
Negative samples are constructed from recovered faces $\mIh_r$ and their ground-truth attributes $\va$ as well as real FR faces and mismatched  (fake) attributes $\tilde{\va}$. 
Thus, the negative sample pairs consist of both $(\mIh_r, \va)$ and $(\mI_r, \tilde{\va})$. 
The parameters of the discriminator $\mPhi$ are updated by minimizing the loss:
%
{
\vspace{-0.18cm}
\begin{equation}
\label{eqn:disc}
\begin{split}
\hspace{-1mm}\mathcal{L}_{dis}(\mPhi)\!= &\!-\!\mathbb{E}_{(\mI_r,\va)\sim p(\mI_r,\va)}[\log \mD_\mathcal{\mPhi}(\mI_r,\va)]
\!\!\\
&-\!\mathbb{E}_{(\mIh_r,\va)\sim p(\mIh_r,\va)}[\log(1\!-\!\mD_\mathcal{\mPhi}(\mIh_r,\va))]\\
\!\!&-\!\mathbb{E}_{(\mI_r,\tilde{\va})\sim p(\mI_r,\tilde{\va})}[log(1\!-\!\mD_\mathcal{\mPhi}(\mI_r,\tilde{\va}))], 
\end{split}\!\!\!
\vspace{-3mm}
\end{equation}
}%
where $p(\mI_r,\va)$, $p(\mIh_r,\va)$ and $p(\mI_r,\tilde{\va})$ denote distributions of real and recovered faces and the corresponding attributes respectively, and $p(\mI_r,\tilde{\va})$ represents the distribution of the recovered faces and the corresponding mismatched (fake) attributes. $\mD_\mPhi(\mI_r,\va)$, $\mD_\mPhi(\mIh_r,\va)$ and $\mD_\mPhi(\mI_r,\tilde{\va})$ are the outputs of $\mD_\mPhi$. 
We update parameters of the discr. network, $\mathcal{L}_{dis}$ loss is back-propagated to FRN. 

Our FNR loss is a weighted sum of three terms: the pixel-wise loss, the discriminative loss, and the identity-preserving loss. The parameters $\mTheta$ are obtained by minimizing the objective function of the FRN loss as follows:
\vspace*{-0.1in}
{
\vspace{-0.2cm}
\begin{align}
&\!\!\!\!\fontsize{9}{9}\selectfont\mathcal{L}_{FNR}(\mTheta)=\!\mathbb{E}_{(\mI_p,\mI_r,\va)\sim p(\mI_p,\mI_r,\va)} \|\mG_{\mTheta}(\mI_p) - \mI_r\|_F^2\nonumber\\
&\quad\fontsize{9}{9}\selectfont\!+\!\lambda\ \mathbb{E}_{\mI_p\sim p(\mI_p,\va))}[\log\!\mD_\mPhi(\mG_{\mTheta}(\mI_p,\va),\va)]\\
&\quad\fontsize{9}{9}\selectfont+\!\eta\ \mathbb{E}_{(\mI_p,\mI_r,\va)\sim p(\mI_p,\mI_r,\va)}\|\vpsi(\mG_{\mTheta}(\mI_p,\va))-\vpsi(\mI_r)\|^2_{F},\nonumber
\vspace{-2mm}
\end{align}
}
where $\lambda$ determines a trade-off between the appearance and the attribute similarity, and $\eta$ determines a trade-off between the image intensity and the feature similarity. 

As $\mG_{\mTheta}(\cdot)$ and $\mD_{\mPhi}(\cdot)$ are differentiable, we apply back-propagation with respect to $\mTheta$ and $\mPhi$, and optimize using Stochastic Gradient Descent (SGD) combined with the Root Mean Square Propagation (RMSprop). %
\vspace{-1mm}
\subsection{Implementation Details}
\vspace{-2mm}
\label{sec:implementation}
The discriminative network $DN$ is needed in the training phase. In the testing phase, we take SP portraits and their corresponding attribute vectors as inputs and feed them to FRN. The outputs of FRN are the recovered photo-realistic face images. Although the attributes used for training are normalized between 0 and 1, they can be scaled up and down, \eg, above 1 or below 0, to manipulate the final results according to users' requirements. 

We use convolutional layers with kernels of size $4\times4$ and stride $2$ in the encoder and deconvolutional layers with kernels of size $4\times4$ and stride $2$ in the decoder. 
We use mini-batches of size 64, the learning rate $10^{-3}$ and the decay rate  $10^{-2}$. For STNs, we  use architectures 
as in~\cite{shiri2018identity}. 
In all  experiments,  $\lambda$ and $\eta$ are set to $10^{-2}$ and $10^{-3}$, respectively, gradually reducing $\lambda$ by a factor $0.995$ to emphasize the importance of appearance similarity. 
As our method is feed-forward at the test time, 
it takes 8 ms to destylize a 128$\times$128 image. 

\vspace{-2mm}
\section{ Dataset and Preprocessing}
\vspace{-2mm}
\label{Sec:Data}
To avoid overfitting in AFRP, a large number of SP/RF training image pairs are obtained from the CelebA dataset \cite{Liu2015faceattributes}. 
We randomly select 110K real faces for training and 2K images for testing. Then, we crop the central part of each image and resize it to $128\!\times\!128$ pixels as our RF ground-truth face images $\mI_r$. We augment RF images by rotation and translation.  
We  use three distinct styles for synthesizing our training dataset 
(see Sec.~\ref{metric}). Finally, we obtain 330K SP/RF pairs and their corresponding attributes for training. We also use 2K unaligned real faces to synthesize 20K SP images from 10 diverse styles as our testing dataset. 
We also add sketches as an unseen style for testing. 
Training and testing datasets are disjoint. %
We choose $20$ dominant attributes (\emph{Bald, Bangs, Big nose, Black Hair, Blond Hair, Brown Hair, Eyeglasses, Gray Hair, Heavy Makeup, Male, Mouth Open, Mustache, Narrow Eyes, No Beard, Pale Skin, Smiling, Straight Hair, Wavy Hair, Wearing Lipstick} and \emph{Young}) from $40$ attributes in CelebA. The ground truth attributes are binary 0/1 values. 
\vspace{-1mm}
\subsection{Style Distance Metric}
\vspace{-2mm}
\label{metric}
It is not practical to generate a large number of styles for training. Thus, we propose a style distance metric to select the most difficult styles for the face recovery process. 
To this end, we compute Gram matrices for various styles from feature maps of pre-trained VGG-network~\cite{simonyan2014very}. Then, we measure the similarity of styles via the Log-Euclidean metric~\cite{jayasumana2013kernel} between Gram matrices of style images and the average Gram matrix of  real training face images. 
As a result, we choose \emph{Candy, Wave} and \emph{Mosaic} styles for training. 

\vspace{-2mm}
\section{Experiments}
\vspace{-2mm}
\label{Exp}

We compare our approach qualitatively/quantitatively to the state-of-the-art methods~\cite{johnson2016perceptual,shiri2017face,isola2016image,zhu2017unpaired,shiri2018identity}. For fairness, we retrain these methods on our training dataset for the task of photorealistic face recovery from stylized portraits.


\begin{figure*}[t]
\vspace{-0.4cm}
\centering
\begin{minipage}{1\linewidth}
\centering
\subfigure[Gender]{\label{fig:attrGT}\includegraphics[width=0.32\linewidth]{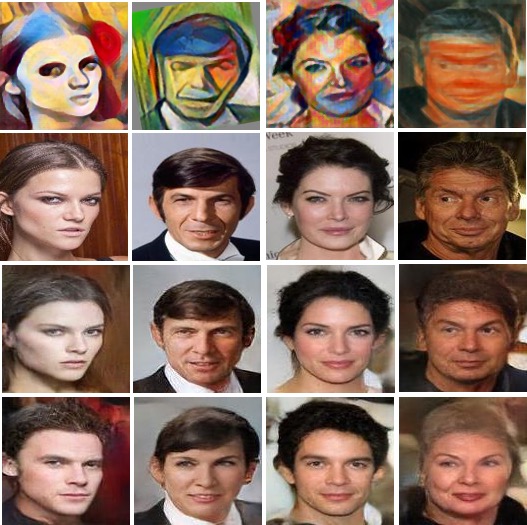}}
\hfill
\subfigure[Age]{\label{fig:attrAge}\includegraphics[width=0.32\linewidth]{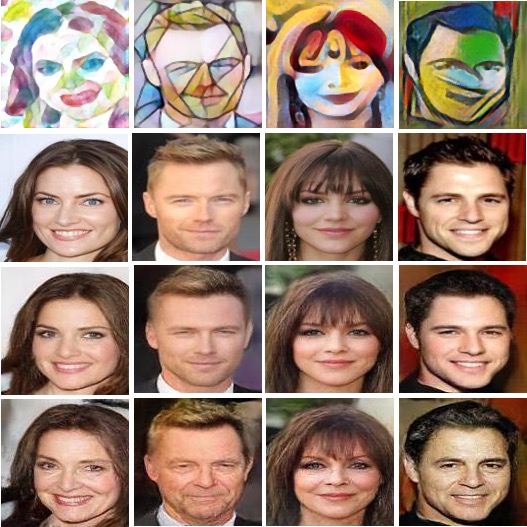}}
\hfill
\subfigure[Makeup]{\label{fig:attrMU}\includegraphics[width=0.32\linewidth]{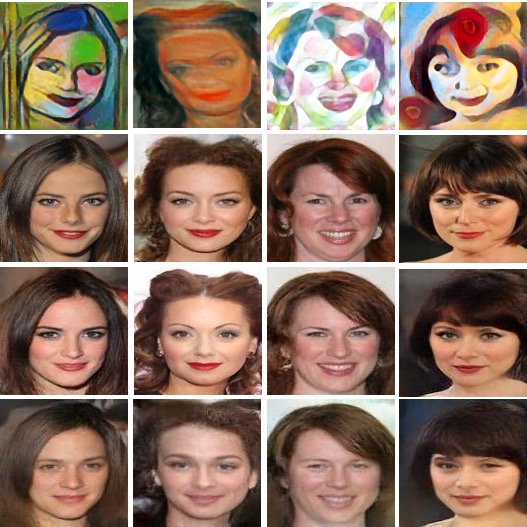}}
\vspace{-0.5em}
\end{minipage}
\vspace{-1mm}
\begin{minipage}{1\linewidth}
\centering
\subfigure[Mouth]{\label{fig:attrMouth}\includegraphics[width=0.32\linewidth]{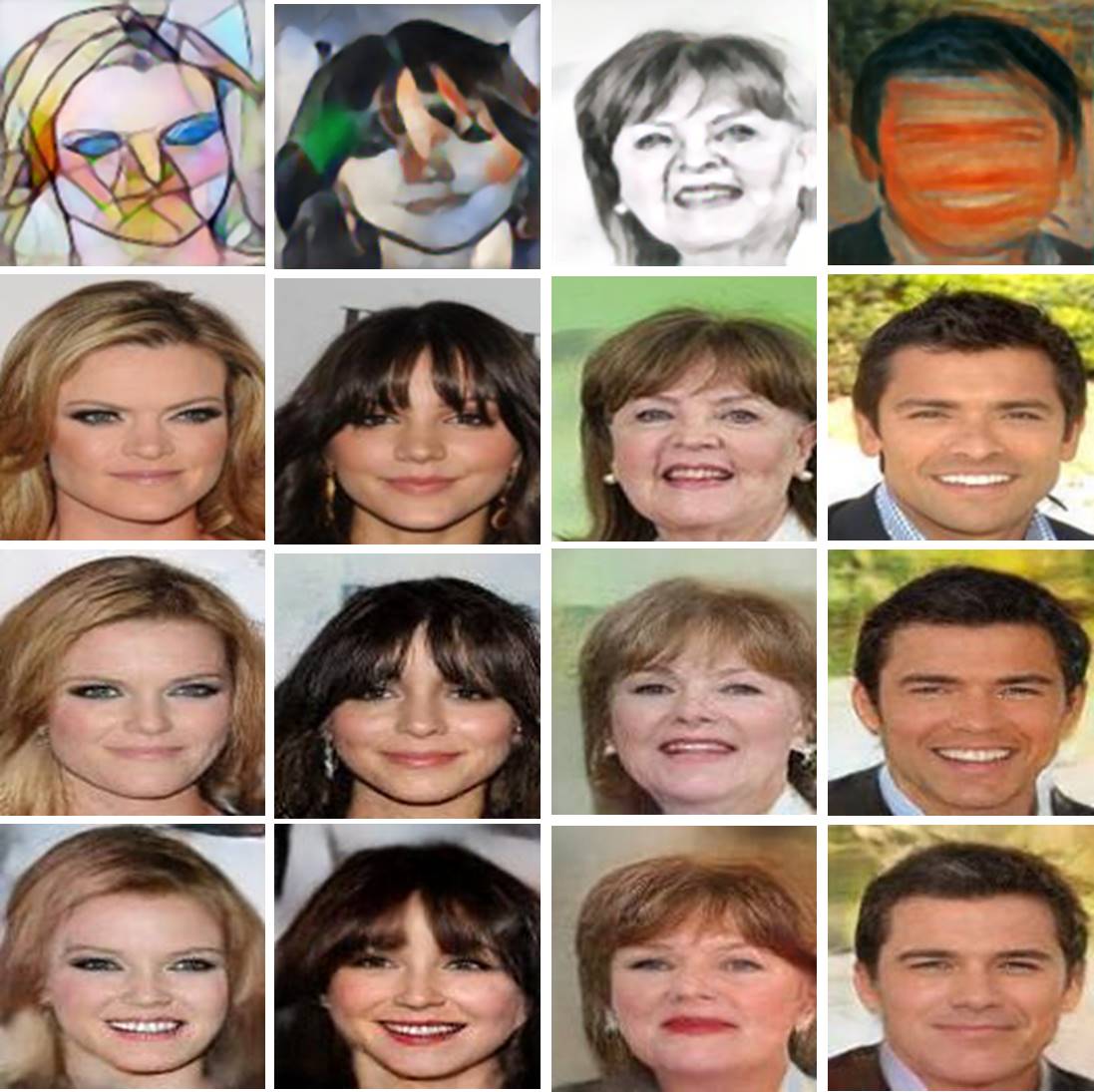}}
\hfill
\subfigure[Beard]{\label{fig:attrBeard}\includegraphics[width=0.32\linewidth]{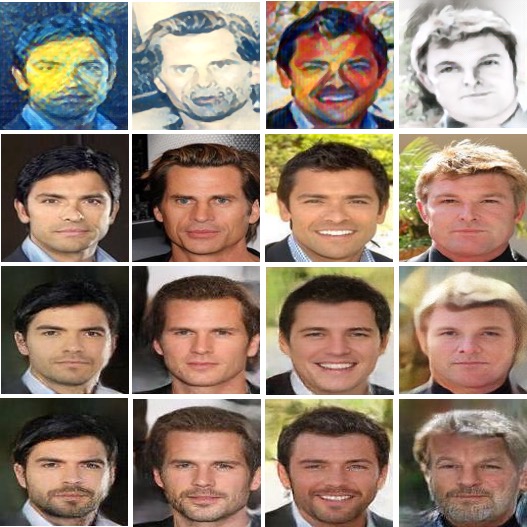}}
\hfill
\subfigure[Hair Color]{\label{fig:attrHC}\includegraphics[width=0.32\linewidth]{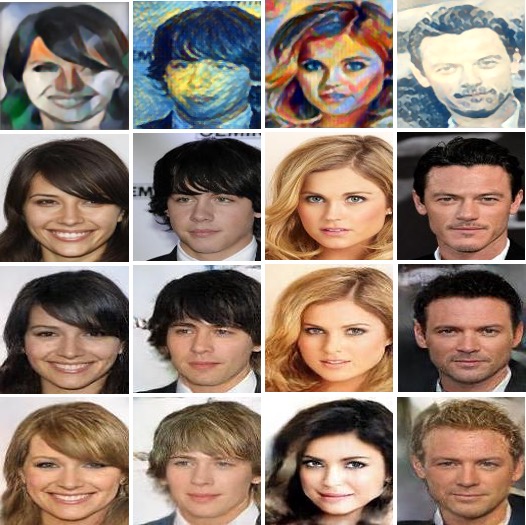}}
\end{minipage}
\vspace{-0.2em}
\caption{Our method lets us fine-tune the recovered results by manipulating the attributes. First row: Unaligned input portraits. Second row: RF ground-truth faces. Third row: Our results with ground-truth attributes. Fourth row: Our results by adjusting attributes. (a) Changing gender. (b) Adding age. (c) Removing makeup. (d) Opening and closing mouth. (e) Adding beard. (d) Changing hair color.}
\label{fig:attr}
\vspace{-1.0em}
\end{figure*}
\vspace{-1mm}
\subsection{Attribute Manipulation in Face Recovery}
\vspace{-1mm}
By manipulating  attribute vectors, we can also post-edit results.
As shown in Fig.~\ref{fig:attrHC}, by changing the hair color attribute, we can restore the same person with different hair color. Our method can also manipulate the age 
 in Fig.~\ref{fig:attrAge}, remove the eye-lines and lipstick in Fig.~\ref{fig:attrMU}, open or close mouths in Fig.~\ref{fig:attrMouth}, add beard in Fig.~\ref{fig:attrBeard}, and change the hair color in Fig.~\ref{fig:attrHC}.
%
\begin{table}[t]\renewcommand{\arraystretch}{1.0}
\vspace{-0.4cm}
\caption{Impact of tuning attributes on the classification results.}
\vspace{-0.2em}
\centering
\setlength{\tabcolsep}{0.10em}
\renewcommand{\arraystretch}{0.70}
\begin{tabular}{>{\centering\arraybackslash}m{0.16\linewidth}|>{\centering\arraybackslash}m{0.22\linewidth}|>{\centering\arraybackslash}m{0.22\linewidth}|>{\centering\arraybackslash}m{0.22\linewidth}}
\hline
Attributes & GT Attr. Acc. & Increased Attr. Acc.  & Decreased Attr. Acc. \\
\hline
\emph{Young} & 95\% & 100\% & 0.5\%\\
\hline
\emph{Male} & 100\% & 100\% & 1\%\\
\hline
\emph{Beard} & 79\% & 100\% & 15\%\\
\hline
\end{tabular}
\label{tab1}
\vspace{-0.5em}
\end{table}

To test if the attribute information has been successfully embedded in our network, we choose three different attributes,~\ie \emph{Young}, \emph{Male} and \emph{Beard}, and we train an attribute classifier for each attribute. By increasing and decreasing the corresponding attribute values, the true positive accuracies change accordingly, as shown in Table~\ref{tab1}. This indicates that the attribute information and thus semantic information has been successfully injected into recovery. 

\vspace{-1mm}
\subsection{Qualitative Evaluation}
\begin{figure*}[t]
\vspace{-0.4cm}
\begin{minipage}{0.1\linewidth}
\centering
\subfigure[RF]{\label{fig:cmp1r}\scalebox{1}[1]
{\includegraphics[width=1.18\linewidth]{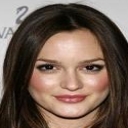}}} 
\end{minipage}
\begin{minipage}{1\linewidth}
\centering
\hspace{-5em}\includegraphics[width=0.118\linewidth]{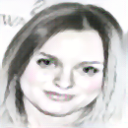}
\includegraphics[width=0.118\linewidth]{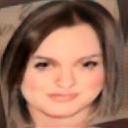}
\includegraphics[width=0.118\linewidth]{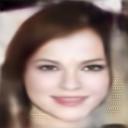}
\includegraphics[width=0.118\linewidth]{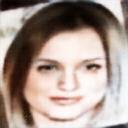}
\includegraphics[width=0.118\linewidth]{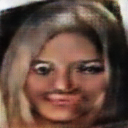}
\includegraphics[width=0.118\linewidth]{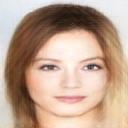}
\includegraphics[width=0.118\linewidth]{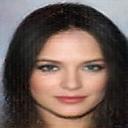}\\\vspace{0.1em}
\hspace{-5em}\includegraphics[width=0.118\linewidth]{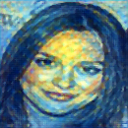}
\includegraphics[width=0.118\linewidth]{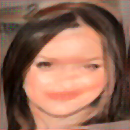}
\includegraphics[width=0.118\linewidth]{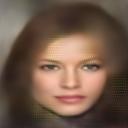}
\includegraphics[width=0.118\linewidth]{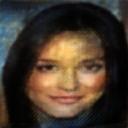}
\includegraphics[width=0.118\linewidth]{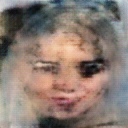}
\includegraphics[width=0.118\linewidth]{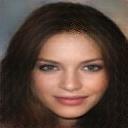}
\includegraphics[width=0.118\linewidth]{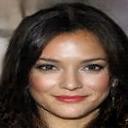}\\\vspace{0.1em}
\hspace{-5em}\includegraphics[width=0.118\linewidth]{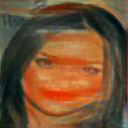}
\includegraphics[width=0.118\linewidth]{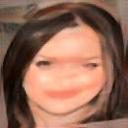}
\includegraphics[width=0.118\linewidth]{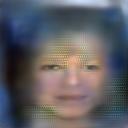}
\includegraphics[width=0.118\linewidth]{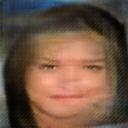}
\includegraphics[width=0.118\linewidth]{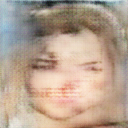}
\includegraphics[width=0.118\linewidth]{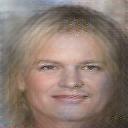}
\includegraphics[width=0.118\linewidth]{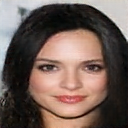}\\\vspace{0.1em}
\hspace{-5em}\includegraphics[width=0.118\linewidth]{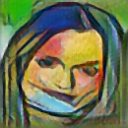}
\includegraphics[width=0.118\linewidth]{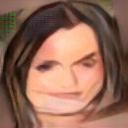}
\includegraphics[width=0.118\linewidth]{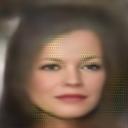}
\includegraphics[width=0.118\linewidth]{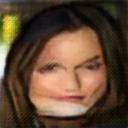}
\includegraphics[width=0.118\linewidth]{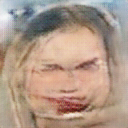}
\includegraphics[width=0.118\linewidth]{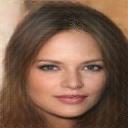}
\includegraphics[width=0.118\linewidth]{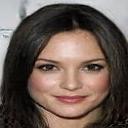}\\\vspace{0.1em}
\hspace{-5em}\includegraphics[width=0.118\linewidth]{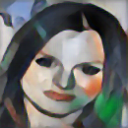}
\includegraphics[width=0.118\linewidth]{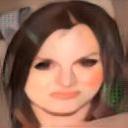}
\includegraphics[width=0.118\linewidth]{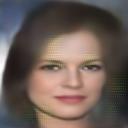}
\includegraphics[width=0.118\linewidth]{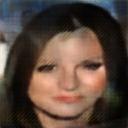}
\includegraphics[width=0.118\linewidth]{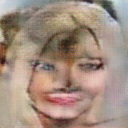}
\includegraphics[width=0.118\linewidth]{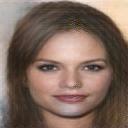}
\includegraphics[width=0.118\linewidth]{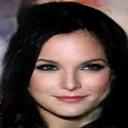}\\\vspace{0.1em}
\hspace{-5em}\includegraphics[width=0.118\linewidth]{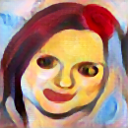}
\includegraphics[width=0.118\linewidth]{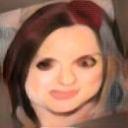}
\includegraphics[width=0.118\linewidth]{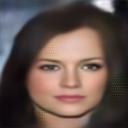}
\includegraphics[width=0.118\linewidth]{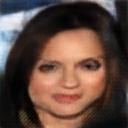}
\includegraphics[width=0.118\linewidth]{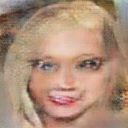}
\includegraphics[width=0.118\linewidth]{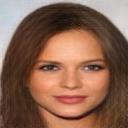}
\includegraphics[width=0.118\linewidth]{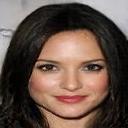}\\\vspace{-0.35em}
\hspace{-5em}\subfigure[SP]{\label{fig:cmp1b}\scalebox{1}[1]{\includegraphics[width=0.118\linewidth]{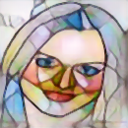}}}
\subfigure[\cite{johnson2016perceptual}]{\label{fig:cmp1c}\scalebox{1}[1]
{\includegraphics[width=0.118\linewidth]{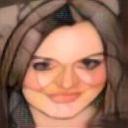}}}
\subfigure[\cite{shiri2017face}]{\label{fig:cmp1d}\scalebox{1}[1]
{\includegraphics[width=0.118\linewidth]{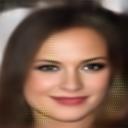}}}
\subfigure[\cite{isola2016image}]{\label{fig:cmp1e}\scalebox{1}[1]
{\includegraphics[width=0.118\linewidth]{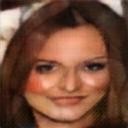}}}
\subfigure[\cite{zhu2017unpaired}]{\label{fig:cmp1f}\scalebox{1}[1]
{\includegraphics[width=0.118\linewidth]{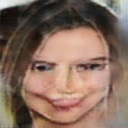}}}
\subfigure[\cite{shiri2018identity}]{\label{fig:cmp1g}\scalebox{1}[1]
{\includegraphics[width=0.118\linewidth]{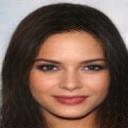}}}
\subfigure[Ours]{\label{fig:cmp1h}\scalebox{1}[1]
{\includegraphics[width=0.118\linewidth]{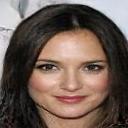}}}
\end{minipage}
\hfill
\vspace{-0.2cm}
\caption{Comparisons to the state of the art. (a) Original RF image. (b) Input portraits (from the test set) including the unseen styles \emph{Sketch, Starry, Scream, La Muse} and \emph{Udnie} and the seen styles \emph{Candy} and \emph{Mosaic}. (c) Johnson \emph{et al.}'s method~\cite{johnson2016perceptual}. (d) Shiri \emph{et al.}'s method~\cite{shiri2017face} (e) Isola \emph{et al.}'s method~\cite{isola2016image} (pix2pix). (f) Zhu \emph{et al.}'s method~\cite{zhu2017unpaired} (CycleGAN). (g) Shiri~\etal's method~\cite{shiri2018identity}. (h) Our method.}
\label{fig:cmp1}
\vspace{-0.2em}
\end{figure*}

\vspace{-0.45em}
We provide sample results in Fig.~\ref{fig:cmp1} (also see our supp. material). Note that \cite{shiri2017face,isola2016image,johnson2016perceptual,zhu2017unpaired} require input SP faces to be aligned before recovery so we employ an STN to align all the SP images. Our method and \cite{shiri2018identity} automatically generate upright real face images. The aligned upright RF ground-truth images are shown for comparison. 
We visually compare our approach with five methods detailed below. 

Johnson \emph{et al.} \cite{johnson2016perceptual} captures  correlation between feature maps of the portrait and the synthesized face (Gram matrices). 
We retrain this approach for destylization. However, it fails to preserve spatial structures of face images and network generates distorted facial details and unnatural artefacts. As shown in Fig.~\ref{fig:cmp1c}, the facial details are blurred and the artistic styles have been removed only partially.

Shiri \emph{et al.} \cite{shiri2017face} introduce a face destylization  which uses only  a pixel-wise loss in their generative network and a standard discriminator to enhance facial details. Although their approach is trained on a large-scale dataset, it fails to generate authentic facial details due to the existence of various styles. As seen in Fig.~\ref{fig:cmp1d}, it produces distorted results and the facial colors are inconsistent. It cannot recover faces from unaligned portraits or large pose portraits. 

Isola~\emph{et al.}~\cite{isola2016image} train a "U-net" generator augmented with a PatchGAN discriminator in an adversarial framework, known as "pix2pix". 
Their network does not capture the global structure of faces. 
As shown in Fig.~\ref{fig:cmp1e},  pix2pix can generate acceptable results for the seen styles but it fails to remove the unseen styles and produces obvious artefacts.

CycleGAN~\cite{zhu2017unpaired} is an image-to-image translation method that uses unpaired datasets. 
Since CycleGAN also employs a patch-based discriminator, it cannot capture the global structure of faces either. As CycleGAN uses unpaired face datasets, 
the low-level features of the stylized faces and real faces do not match well.
As shown in Fig.~\ref{fig:cmp1f}, this method produces distorted results and does not preserve the identities with respect to the input images.

Shiri \emph{et al.} \cite{shiri2018identity} exploit an identity-preserving loss to reveal the photorealistic faces from unaligned stylized faces. 
They also employ a simple autoencoder and standard discriminative network to recover the real faces, but their discriminative network is only used to force the generative network to produce sharper results without imposing attribute information. As shown in Fig.~\ref{fig:cmp1g}, their method suffers mismatched hair colors. As shown in the third row of Fig.~\ref{fig:cmp1g}, their method also recovers male facial details. 

In contrast, our results demonstrate higher fidelity and better consistency with respect to the ground-truth face images as shown in Fig.~\ref{fig:cmp1h}. Our method, evaluated on portraits from seen/unseen styles and sketches, produces high-quality realistic faces which also match the semantic composition of ground-truth images. 
Our network recovers the photorealistic faces from various stylized portraits of the same person as shown in Fig.~\ref{fig:cmp1}. Note that the recovered faces resemble each other. This demonstrates the robustness of our network with respect to different styles. 

\vspace{-1mm}
\subsection{Quantitative Evaluation}
\vspace{-1mm}
\noindent{\textbf{ Face Reconstruction Analysis. }}
To evaluate the reconstruction performance, we measure the average Peak Signal to Noise Ratio (PSNR) and Structural Similarity (SSIM)~\cite{wang2004image}  on the entire test dataset. Table~\ref{tab2} indicates that our method achieves superior quantitative performance in comparison to other methods on both seen and unseen styles. 
As indicated in Table~\ref{tab2}, we show the quantitative results of solely using FRN, marked as FRN. Also, the results of using both FRN and a standard DN, indicated by FRN+SDN, is demonstrated in Table~\ref{tab2}.
The standard DN only forces FRN to generate realistic faces, and thus it improves the results qualitatively and quantitatively.
Since FRN augmented with attributes may learn a trivial solution, where all attribute vectors will be neglected, using a standard DN cannot force FRN to embed such attribute information. 
On the contrary, our conditional DN is able to distinguish whether the attributes match the input faces or not, thus forcing FRN to embed attribute information in the process of face recovery. Thus, the ambiguity is significantly reduced and the network achieves better performance.

\noindent\textbf{Face Retrieval Analysis. }
To demonstrate that the faces recovered by our method are highly consistent with their ground-truth counterparts,
we run a face recognition algorithm~\cite{parkhi2015deep} on our test dataset for both seen and unseen styles. For each investigated method, we consider 2K recovered faces from one style as query images and then use their ground-truth real faces as a gallery dataset. 
We run \cite{parkhi2015deep} to check whether the correct person is retrieved within the top-5 matched images and then an average retrieval score is obtained. We repeat this procedure for each style and then obtain the average Face Retrieval Ratio (FRR) by averaging all scores from the seen and unseen styles, respectively. 
As indicated in Table~\ref{tab3}, our AFRP network outperforms the other methods across all the styles. Even for the unseen styles, our method can still generate realistic facial details with high fidelity to the ground-truth.
\begin{figure*}[t]
\vspace{-0.2em}
\centering
\subfigure[]{\includegraphics[width=0.18\linewidth]{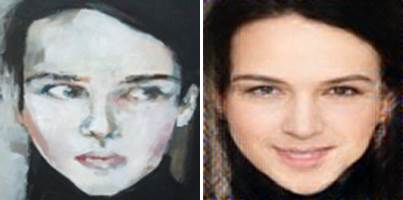}}
\subfigure[]{\includegraphics[width=0.18\linewidth]{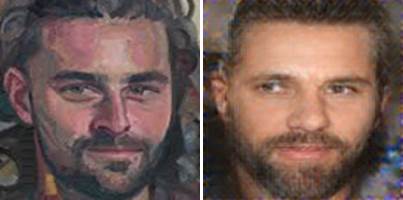}}
\subfigure[]{\includegraphics[width=0.2\linewidth]{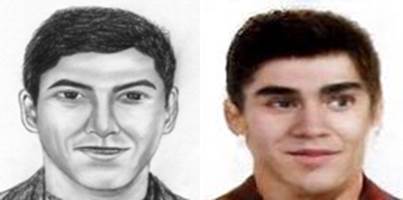}}
\subfigure[]{\includegraphics[width=0.2\linewidth]{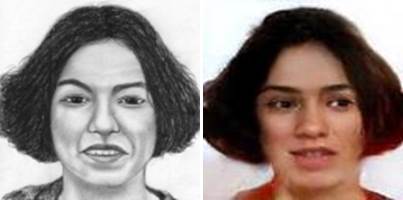}}

\vspace{-0.1em}
\caption{Results for the original unaligned paintings and hand-drawn sketches. Right: the original portraits. Left: our results.}
\label{fig:Orig}
\end{figure*}

\begin{table}[!t]\renewcommand{\arraystretch}{1}
\centering
\setlength{\tabcolsep}{0.10em}
\renewcommand{\arraystretch}{0.70}
\vspace{-0.1cm}
\caption{Comparisons of PSNR and SSIM on the entire test dataset.}
\vspace{-0.05em}
\scalebox{0.9}{
\begin{tabular}{>{\centering\arraybackslash}m{0.3\linewidth}|>{\centering\arraybackslash}m{0.12\linewidth}|>{\centering\arraybackslash}m{0.12\linewidth}|>{\centering\arraybackslash}m{0.12\linewidth}|>{\centering\arraybackslash}m{0.12\linewidth}|>{\centering\arraybackslash}m{0.12\linewidth}|>{\centering\arraybackslash}m{0.07\linewidth}}
\hline
\multirow{2}{*}{Method} & \multicolumn{2}{c|}{Seen Styles} & \multicolumn{2}{c|}{Unseen Styles} & \multicolumn{2}{c}{Unseen Sketches} \\
\cline{2-7}
& PSNR & SSIM & PSNR & SSIM  & PSNR & SSIM \\
\hline
Johnson~\cite{johnson2016perceptual} & 17.85 & 0.76 & 18.07 & 0.75 & 18.11 & 0.76\\ 
Shiri~\cite{shiri2017face} & 19.22 & 0.81 & 19.09 & 0.80 & 19.01 & 0.80\\ 
pix2pix~\cite{isola2016image} & 18.45 & 0.78 & 18.12 & 0.77 & 18.07 & 0.78\\ 
CycleGAN~\cite{zhu2017unpaired} & 18.35 & 0.75 & 18.29 & 0.75 & 18.08 & 0.76\\ 
Shiri~\cite{shiri2018identity} & 19.35 & 0.80 & 19.31 & 0.79 & 19.045 & 0.80\\
\hline
FRN & 18.42 & 0.76 & 18.51 & 0.75 & 18.49 & 0.75 \\
FRN + SDN & 19.66 & 0.81 & 19.58 & 0.80 & 19.62 & 0.80 \\
\bf{AFRP} & {\bf 20.01} & {\bf 0.84} & {\bf 19.99} & {\bf 0.83} & {\bf 19.98} & {\bf 0.83}\\
\hline
\end{tabular}}
\label{tab2}
\vspace{-1.2em}
\end{table}
\vspace{-1mm}
\subsection{Destylizing Original Paintings and Sketches}
\vspace{-2mm}
Fig.~\ref{fig:Orig} illustrates that our method is not limited to computer-generated stylized portraits and it can also efficiently recover photorealistic faces from original paintings and sketches.
We choose real paintings from art galleries and hand-drawn sketches as our test examples. Since we do not know the ground-truth attributes, we set the attribute vectors to neutral values \ie, 0.5. As shown in Fig.~\ref{fig:Orig}, despite attributes may be inaccurate, our method still  generates authentic face images regardless of their original styles.
\begin{table}[!t]\renewcommand{\arraystretch}{1}
\centering
\caption{Comparisons of FRR on the entire test dataset.}
\scalebox{0.9}{
\begin{tabular}{>{\centering\arraybackslash}m{0.245\linewidth}|>{\centering\arraybackslash}m{0.20\linewidth}|>{\centering\arraybackslash}m{0.24\linewidth}|>{\centering\arraybackslash}m{0.26\linewidth}}
\hline
Method & Seen Styles & Unseen Styles & Unseen Sketch \\
\hline
Johnson~\cite{johnson2016perceptual} & 55.57\%  & 50.48\% & 54.36\% \\ 
Shiri~\cite{shiri2017face} & 78.00\%  & 66.89\% & 65.26\% \\ 
pix2pix~\cite{isola2016image} & 76.03\%  & 62.67\% & 64.64\% \\ 
CycleGAN~\cite{zhu2017unpaired} & 36.07\% &  33.68\% & 32.75\% \\ 
Shiri~\cite{shiri2018identity} & 84.51\%  & 75.32\% & 75.44\% \\ 
\hline
AFRP & {\bf 93.08\%} &  {\bf 83.14\%} & {\bf 92.05\%}  \\
\hline
\end{tabular}}
\label{tab3}
\end{table}

\vspace{-0.3cm}
\section{Conclusions}
\vspace{-1mm}
We have introduced an attribute guided generative-discriminative network to recover photorealistic faces from unaligned stylized portraits in an end-to-end fashion. With the help of the conditional discriminative network, our network successfully incorporates facial attribute vectors into the residual features of input portraits at the bottleneck of the autoencoder.
Our network is able to preserve the identity of generated faces and it can post-edit the recovered results by adjusting the attribute information. Moreover, our algorithm demonstrates good generalization ability for recovery of portraits from unseen styles, real paintings as well as hand-drawn sketches.

\vspace{0.1cm}
\noindent{\textbf{Acknowledgement. }}
This work is supported by the Australian Research Council (ARC) grant DP150104645.

\renewcommand*\appendixpagename{\fontsize{15}{16}\selectfont Appendices\vspace{-0.3cm}}
\begin{appendices}


\section{Synthesized Dataset}
\begin{figure}
\centering
\subfigure[]{\label{fig:dataset1}\scalebox{1}[1]{\includegraphics[width=0.15\linewidth]{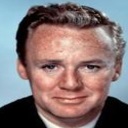}}}
\subfigure[]{\label{fig:dataseta}\scalebox{1}[1]{\includegraphics[width=0.15\linewidth]{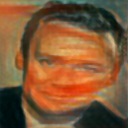}}}
\subfigure[]{\label{fig:datasetb}\scalebox{1}[1]{\includegraphics[width=0.15\linewidth]{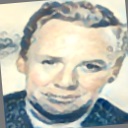}}}
\subfigure[]{\label{fig:datasetc}\scalebox{1}[1]{\includegraphics[width=0.15\linewidth]{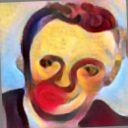}}}
\subfigure[]{\label{fig:datasetd}\scalebox{1}[1]{\includegraphics[width=0.15\linewidth]{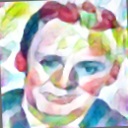}}}\\

\subfigure[]{\label{fig:datasete}\scalebox{1}[1]{\includegraphics[width=0.15\linewidth]{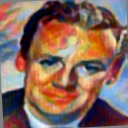}}}
\subfigure[]{\label{fig:datasetf}\scalebox{1}[1]{\includegraphics[width=0.15\linewidth]{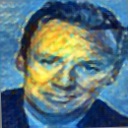}}}
\subfigure[]{\label{fig:datasetg}\scalebox{1}[1]{\includegraphics[width=0.15\linewidth]{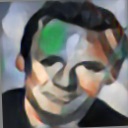}}}
\subfigure[]{\label{fig:dataseth}\scalebox{1}[1]{\includegraphics[width=0.15\linewidth]{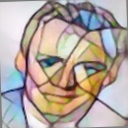}}} 
\subfigure[]{\label{fig:dataseti}\scalebox{1}[1]{\includegraphics[width=0.15\linewidth]{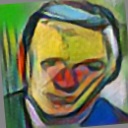}}} 
\subfigure[]{\label{fig:datasetj}\scalebox{1}[1]{\includegraphics[width=0.15\linewidth]{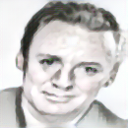}}} \vspace{-0.5em}
\caption{Samples of the synthesized dataset. (a) The ground-truth aligned real face image. (b)-(k) The synthesized unaligned portraits form~\emph{Scream, Wave, Candy, Feathers, Composition VII, Starry night, Udnie, Mosaic,la Muse} and \emph{Sketch} styles which have been used for training and testing our network.}
\label{fig:datasets}
\end{figure}

Figure~\ref{fig:datasets} shows the stylized samples that are generated from a single real image containing a face.
\newcommand{\PowW}{0.14}
\section{Additional Experiments}
\begin{figure*}[!t]
\centering

\subfigure[RF]{\label{fig:cmp1as}\scalebox{1}[1]{\includegraphics[width=0.119\linewidth]{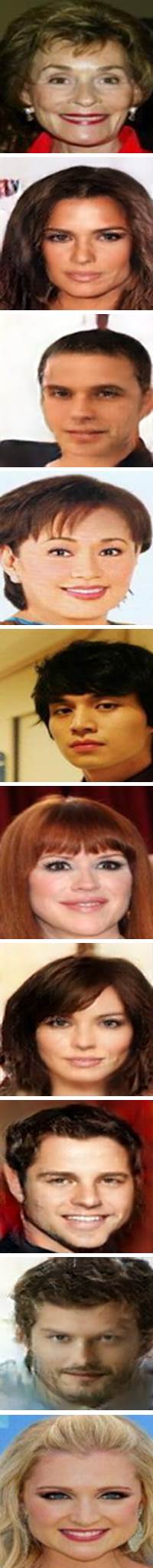}}}
\subfigure[SP]{\label{fig:cmp1bs}\scalebox{1}[1]{\includegraphics[width=0.119\linewidth]{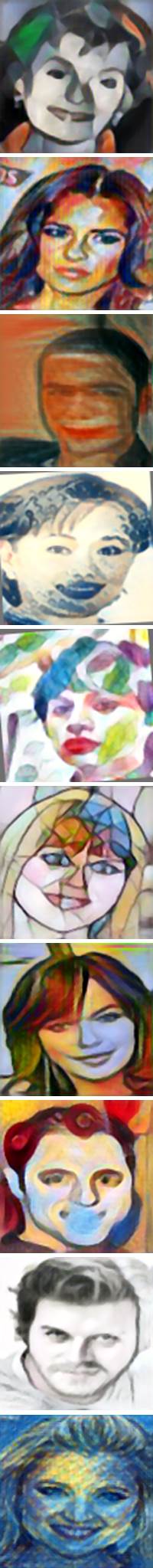}}}
\subfigure[\cite{johnson2016perceptual}]{\label{fig:cmp1cs}\scalebox{1}[1]{\includegraphics[width=0.119\linewidth]{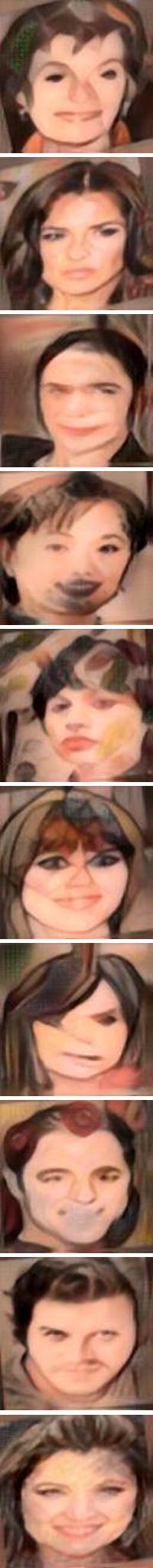}}}
\subfigure[\cite{shiri2017face}]{\label{fig:cmp1ds}\scalebox{1}[1]{\includegraphics[width=0.119\linewidth]{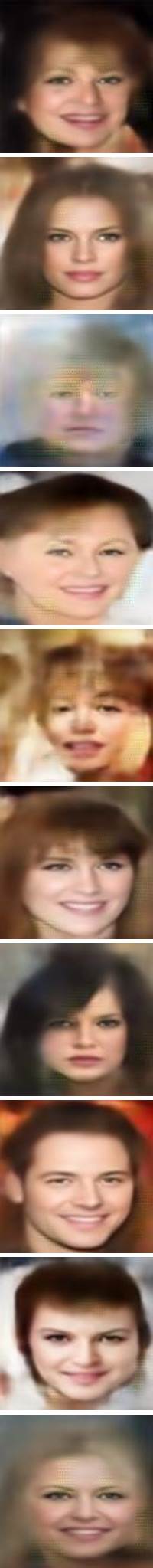}}}
\subfigure[\cite{isola2016image}]{\label{fig:cmp1es}\scalebox{1}[1]{\includegraphics[width=0.119\linewidth]{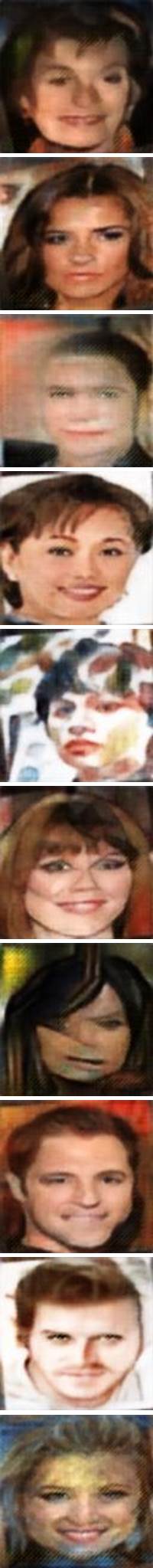}}}
\subfigure[\cite{zhu2017unpaired}]{\label{fig:cmp1fs}\scalebox{1}[1]{\includegraphics[width=0.119\linewidth]{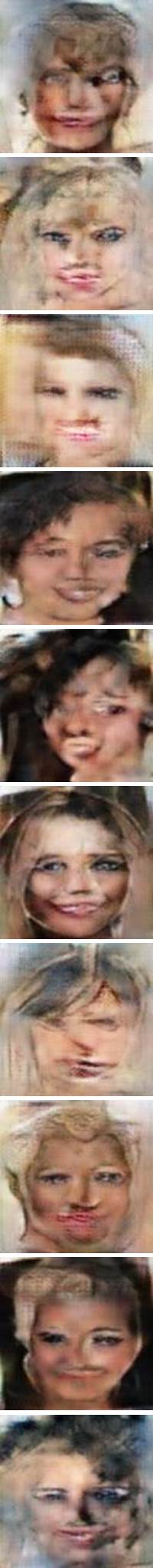}}}
\subfigure[\cite{shiri2018identity}]{\label{fig:cmp1gs}\scalebox{1}[1]{\includegraphics[width=0.119\linewidth]{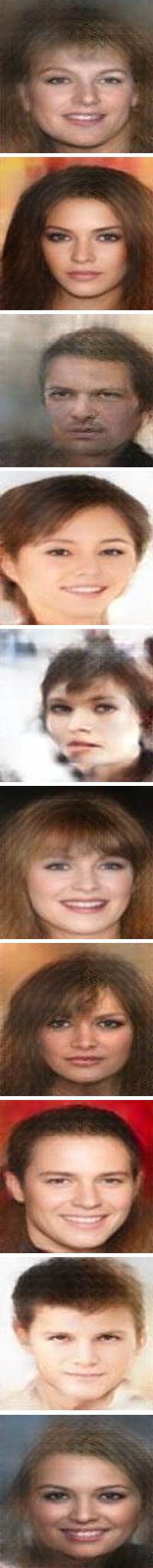}}}
\subfigure[Ours]{\label{fig:cmp1hs}\scalebox{1}[1]{\includegraphics[width=0.119\linewidth]{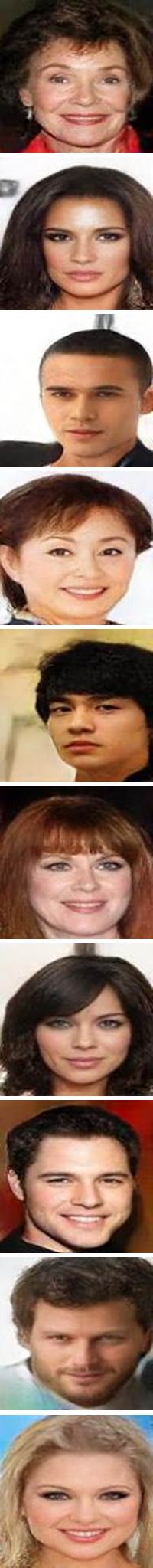}}}

\caption{Comparisons to the state-of-the-art methods. (a) The original RF images. (b) Input portraits (from the test dataset) including the unseen styles as well as the seen styles. (c) Johnson \emph{et al.}'s method~\cite{johnson2016perceptual}. (d) Shiri \emph{et al.}'s method~\cite{shiri2017face} (e) Isola \emph{et al.}'s method~\cite{isola2016image} (pix2pix). (f) Zhu \emph{et al.}'s method~\cite{zhu2017unpaired} (CycleGAN). (g) Shiri~\etal's method~\cite{shiri2018identity}. (h) Our method.}
\label{fig:cmp1s}
\end{figure*}

In Figure \ref{fig:cmp1s} on the next page, we provide additional results demonstrating the performance of our AFRP network compared to the state-of-art approaches.

\end{appendices}

{\small
\bibliographystyle{ieee}
\bibliography{egbib}
}
\nopagebreak
\appendix

\end{document}